\documentclass[letterpaper, 10 pt, conference]{ieeeconf}

\IEEEoverridecommandlockouts
\usepackage{cite}
\usepackage{algorithm}
\usepackage{booktabs}
\usepackage{multirow}
\usepackage{amsmath,amssymb,amsfonts}
\usepackage{algorithmic}
\usepackage{graphicx}
\usepackage{textcomp}
\usepackage{xcolor}
\usepackage{comment}

\def\BibTeX{{\rm B\kern-.05em{\sc i\kern-.025em b}\kern-.08em
    T\kern-.1667em\lower.7ex\hbox{E}\kern-.125emX}}

\title{\LARGE \bf
An Adaptive Differentially Private Federated Learning Framework
}

\author{Jin Wang$^{1}$, Hui Ma$^{1,*}$, Yajun Zhang$^{1}$, Xinjun Pei$^{1}$, Ming Yan$^{1}$, Fei Xing$^{2}$, and Yikun Chen$^{3}$%
\thanks{*Corresponding author: Hui Ma (e-mail: huima@xju.edu.cn).}%
\thanks{$^{1}$Jin Wang, Hui Ma, Yajun Zhang, Xinjun Pei, and Ming Yan are with Xinjiang Key Laboratory of Intelligent Computing and Smart Applications, School of Software, Xinjiang University, Urumqi 830017, China (e-mail: 107552401629@stu.xju.edu.cn; huima@xju.edu.cn; zyj@xju.edu.cn; peixinjun@xju.edu.cn; yanmingtop@gmail.com).}%
\thanks{$^{2}$Fei Xing is with the College of Geography and Remote Sensing Sciences, Xinjiang University, Urumqi, China (e-mail: xingfei@xju.edu.cn).}%
\thanks{$^{3}$Yikun Chen is with Jilin University, Changchun 130012, China.}%
\thanks{This work was supported by the Tianchi Talents - Young Doctor Program (5105250183m), Science and Technology Program of Xinjiang Uyghur Autonomous Region (2024B03028, 2025B04051), Regional Fund of the National Natural Science Foundation of China (202512120005).}%
}

\begin{document}

\maketitle
\thispagestyle{empty}
\pagestyle{empty}

\begin{abstract}
Federated learning enables collaborative model training across distributed clients while preserving data privacy. However, in practical deployments, device heterogeneity and non-independent and identically distributed (Non-IID) data often lead to unstable and biased gradient. When differential privacy is enforced, conventional fixed gradient clipping and Gaussian noise injection may further amplify gradient perturbations, resulting in training oscillation and degraded model performance. To address these challenges, we propose an adaptive differentially private federated learning framework that explicitly targets model efficiency under heterogeneous and privacy-constrained settings. On the client side, a lightweight local dimensionality reduction module is introduced to learn reduced-dimensional intermediate representations and produce more structured gradients during backpropagation, thereby mitigating noise amplification during local optimization. On the server side, an adaptive gradient clipping strategy dynamically adjusts clipping thresholds based on historical update statistics to avoid over-clipping and noise domination. Furthermore, a constraint-aware robust aggregation mechanism is designed to suppress unreliable or noise-dominated client updates and stabilize global optimization. Extensive experiments on CIFAR-10, SVHN, and STL-10 demonstrate that the proposed method consistently improves convergence stability and classification performance under differential privacy.
\end{abstract}


\section{Introduction}
Federated learning (FL) has become a prominent framework for collaborative model training across a set of distributed clients while retaining data locally, thereby facilitating learning in privacy-sensitive and communication-constrained edge scenarios. With the proliferation of mobile devices and Internet-of-Things (IoT) systems\cite{ma2025cellular}, data are inherently generated and stored on end devices, making on-device and cross-device training increasingly important for real-world applications~\cite{byeon2024image,srinivasu2024medical}.

Despite its appeal, FL raises non-trivial privacy and security concerns, since sharing model updates may still leak sensitive information about local data~\cite{hu2025fgsfl}. To provide rigorous privacy guarantees, differential privacy (DP) has become one of the most widely adopted privacy-preserving mechanisms for FL, typically implemented by $\ell_2$-norm clipping followed by Gaussian noise injection~\cite{liu2021federated,kato2024uldpfl}. This DP-SGD style design provides a principled means to bound per-client influence, yet fixed clipping thresholds may fail to track time-varying and client-dependent gradient distributions, leading to over-clipping (signal loss) or under-clipping (noise domination), and hence training oscillation and utility degradation under a fixed privacy budget~\cite{shen2023pldpfldca,shi2025flatterdpfl,andrew2024oneshotprivacyfl}.

Another fundamental challenge in practical FL is rooted in the ubiquitous statistical heterogeneity (Non-IID data) across clients. When local data distributions deviate substantially from the population distribution, the local objectives become misaligned, and the resulting client updates can be highly divergent in both magnitude and direction. Such update disagreement induces client drift and biases the aggregated update direction, which often manifests as slowed convergence, unstable training dynamics, and degraded generalization. Although a large body of prior work has investigated mitigation strategies (e.g., personalization, local regularization/control, and server-side correction or reweighting), robust optimization under realistic Non-IID conditions remains challenging: the distribution-induced bias and directional inconsistency can persist throughout training, and naive averaging may further amplify drift and destabilize global optimization~\cite{liu2025asyncfad}. Consequently, federated models may suffer noticeable performance degradation in heterogeneous deployments.

Motivated by these two challenges, we propose FedDimDP, an adaptive DP-FL framework that enhances model efficiency under Non-IID heterogeneity and DP perturbations through three coordinated designs. First, we introduce a lightweight local dimensionality reduction module to regularize intermediate representations and reduce update variability, which mitigates DP-noise amplification during local training and produces more stable privatized updates. Second, we develop an adaptive DP clipping strategy that calibrates the clipping bound using historical update-norm statistics, so that the bound stays consistent with the evolving update scale and alleviates both signal loss from over-clipping and noise domination from under-clipping. Finally, we design a constraint-aware robust aggregation mechanism that integrates stability- and utility-aware criteria into server aggregation, down-weighting unstable or noise-dominated client updates and stabilizing the global descent direction under DP perturbations and Non-IID drift.

The main contributions of this work are summarized as follows:
\begin{itemize}
    \item \textbf{Differentially private training with adaptive clipping.} We propose an adaptive DP mechanism that calibrates the clipping bound using historical update-norm statistics, improving the utility--privacy trade-off and stabilizing optimization under heterogeneous and time-varying client updates.

    \item \textbf{Constraint-aware robust aggregation.} We introduce a constraint-aware robust aggregation mechanism into differentially private federated learning for image classification, where stability/utility-aware reweighting and a lightweight direction correction are jointly employed to suppress noise-dominated updates and mitigate DP-induced perturbations as well as Non-IID drift.

    \item \textbf{Experimental validation.} Extensive experiments on CIFAR-10, SVHN, and STL-10 demonstrate that the proposed method consistently improves convergence stability and classification performance compared with representative state-of-the-art baselines.
\end{itemize}

\section{Related Work}
\subsection{Differential Privacy in Federated Learning}
Differential privacy (DP) has become a canonical mechanism for protecting sensitive information in federated learning (FL), due to its formal privacy guarantees and broad compatibility with gradient-based optimization~\cite{kamath2024beltandbraces}. In practice, the dominant DP-FL pipeline follows the DP-SGD paradigm, where per-iteration updates are first bounded by $\ell_2$-norm clipping and then perturbed via Gaussian noise~\cite{huang2024sirenpp,xu2024camel}. Despite its simplicity, a key limitation of mainstream DP-FL lies in the widespread use of fixed clipping thresholds, which often fail to accommodate client- and stage-dependent variations of gradient norms in real deployments~\cite{malekmohammadi2024noiseaware}. When the clipping bound is poorly calibrated, the training signal may be excessively truncated (utility loss) or insufficiently controlled (perturbation-dominated updates), both of which can manifest as training oscillation and performance degradation~\cite{hu2024fedsmp,sparsification2025ccs}.

To mitigate this issue, recent studies explore adaptive clipping by exploiting gradient statistics, loss dynamics, or related training signals to update the clipping radius online~\cite{ling2024heterodp}. While adaptive strategies can improve utility in certain regimes, reliably estimating a stable and representative clipping bound remains non-trivial in the presence of strong statistical heterogeneity and fluctuating client behaviors: the observed norms themselves can be noisy and biased, which in turn makes threshold adaptation prone to oscillation or drift~\cite{hu2024fedsmp}. Consequently, developing lightweight yet robust adaptive DP mechanisms that remain stable across heterogeneous clients and training stages is still a practically important open direction.

\subsection{Robust Aggregation under Non-IID Data}
A separate line of research focuses on improving the reliability of server-side aggregation in heterogeneous federated systems. Under Non-IID data, varying local computation, and inconsistent participation, client updates can differ substantially in both magnitude and direction, causing naive averaging (e.g., FedAvg) to exhibit slow or unstable convergence~\cite{kiani2025timeadaptive,wu2023ldpfl,wang2024multiprivacy}. To address this, optimization-oriented approaches incorporate regularization or normalization to better align local training dynamics, including representative methods such as FedProx, FedDyn, and FedNova~\cite{tayyeh2024dpfederated}. These designs aim to reduce client drift and reconcile discrepant local steps through principled modifications of the update rules.

In parallel, robust aggregation methods explicitly target unreliable or anomalous updates by filtering or reweighting client contributions~\cite{vu2024robustdpfl}. Classical Byzantine-resilient rules attempt to suppress extreme deviations but may inadvertently exclude informative minority clients when update distributions are intrinsically skewed under Non-IID data~\cite{wu2023ldpfl}. More recent studies formulate aggregation as a robustness-aware decision problem and assign aggregation weights according to stability indicators or historical consistency, thereby improving robustness against heterogeneous and fluctuating client updates~\cite{wang2024multiprivacy,malekmohammadi2024noiseaware,ling2024heterodp}. Overall, these works highlight the importance of aggregation-level robustness as a complementary axis for stabilizing federated optimization in practical heterogeneous deployments.

\section{Method}
\begin{figure}[!t]
\centering
\includegraphics[width=0.47\textwidth]{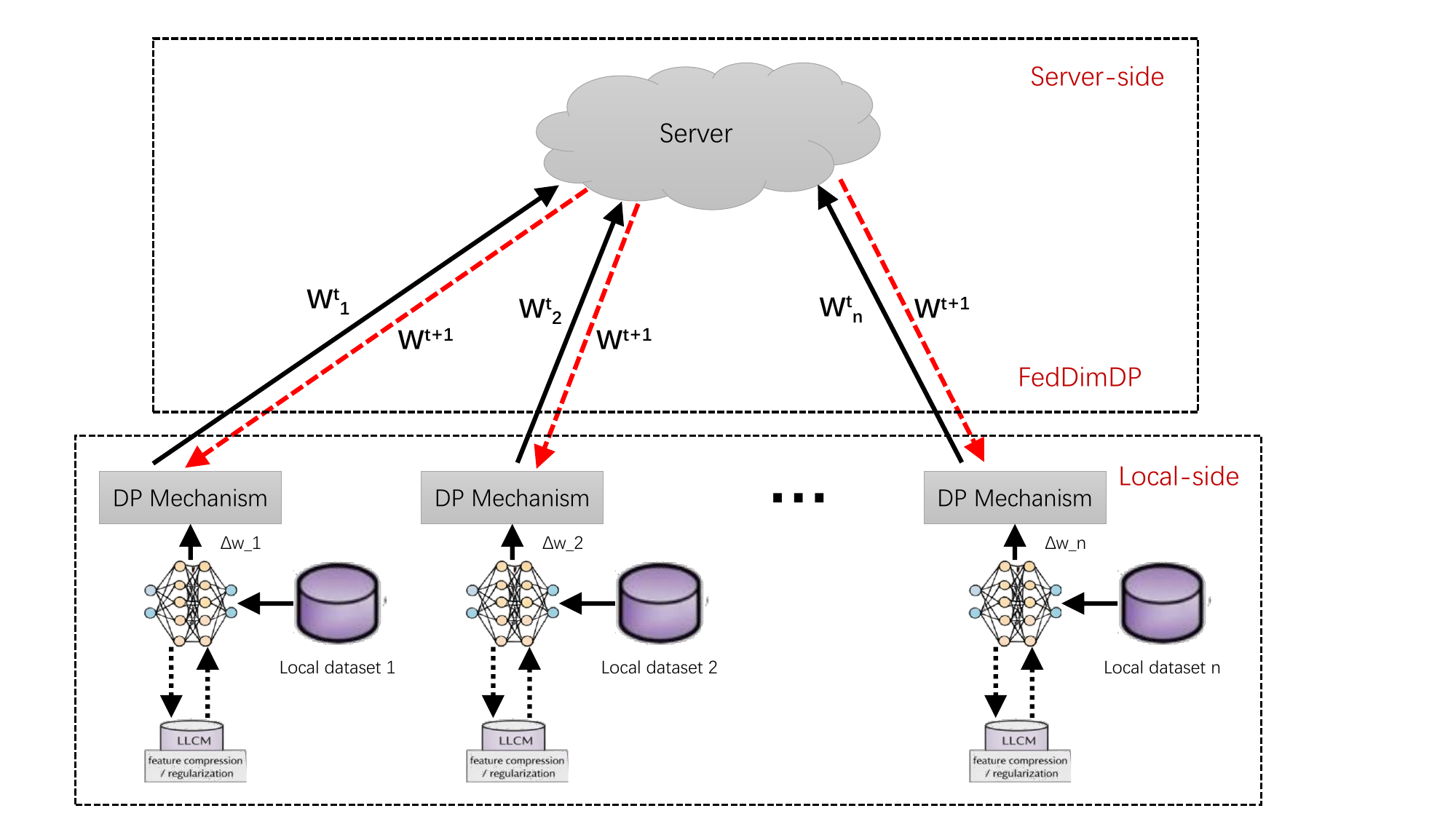}
\caption{Overview of our proposed FedDimDP framework.}
\label{fig1}
\end{figure}
This section presents FedDimDP, an adaptive differentially private federated learning framework tailored to the coupled challenges of Non-IID heterogeneity and privacy leakage risk. As illustrated in Fig.~\ref{fig1}, FedDimDP follows a round-based federated optimization protocol with a central server: in each communication round, all clients perform differentially private local training on their private data and upload privatized model updates; the server then aggregates the collected updates to refine the global model and broadcasts it back for the next round. FedDimDP is built upon three complementary modules that jointly improve model efficiency under heterogeneous data distributions and DP noise. Lightweight Local Dimensionality Reduction Module (LLDM) regularizes intermediate representations via lightweight channel-wise dimensionality reduction, reducing gradient variability during backpropagation and suppressing privacy-noise amplification in local optimization. Adaptive DP Gradient Clipping (ADPC) calibrates the clipping bound using historical update-norm statistics, so that the privacy constraint remains aligned with the evolving scale of client updates and avoids both over-clipping and noise domination. Constraint-aware Robust Aggregation (CRA) incorporates stability- and utility-aware constraints into server aggregation, down-weighting unreliable or noise-dominated updates and stabilizing the global optimization direction under Non-IID drift and DP perturbations.

\subsection{Lightweight Local Dimensionality Reduction Module}
\begin{figure}[t]
    \centering
    \includegraphics[width=\linewidth,trim=0cm 2.0cm 0cm 0.1cm,clip]{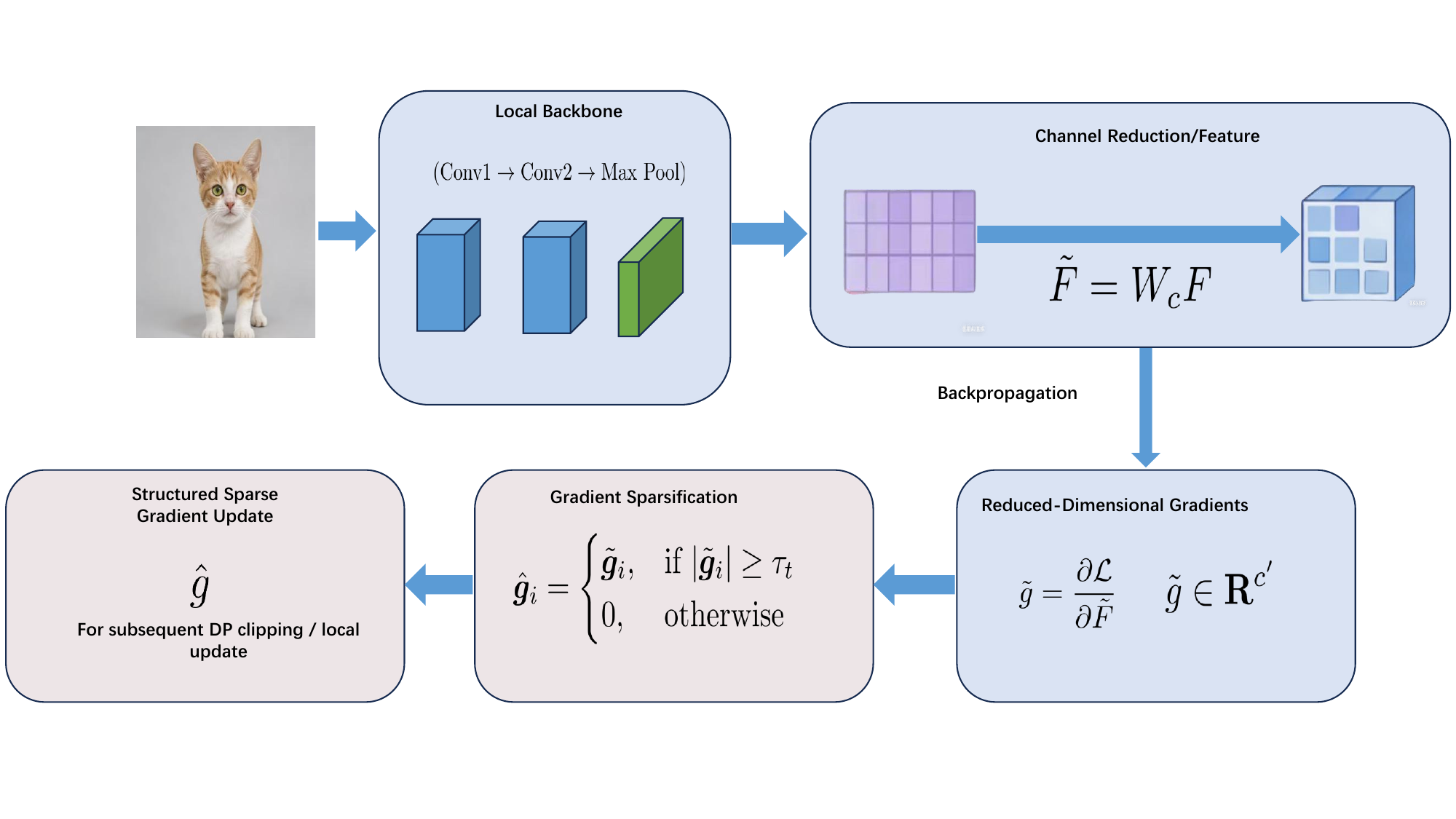}
    \caption{Overview of the proposed lightweight local dimensionality reduction module (LLDM).}
    \label{fig:lldm}
\end{figure}
Under Non-IID data distributions, local training on resource-constrained devices often produces highly variable and noisy gradient representations due to feature imbalance and redundant high-level activations. Such instability not only degrades local optimization efficiency, but also exacerbates the sensitivity of gradient updates to differential privacy noise.

We introduce a lightweight local dimensionality reduction module (LLDM) into the client-side model. As illustrated in Fig.~\ref{fig:lldm}, LLDM consists of channel reduction on intermediate feature representations and gradient sparsification during backpropagation, aiming to impose more structured and stable gradients during local optimization. As a result, the convergence of local training is improved, which in turn enhances the stability of global aggregation in the federated learning process.

\subsubsection{Channel Reduction and Feature Stabilization}
Given an intermediate convolutional feature map $\mathbf{F} \in \mathbb{R}^{C \times H \times W}$, high-dimensional feature representations often contain redundant and highly correlated channels, which may amplify gradient fluctuations under Non-IID data. To mitigate this issue, we apply a linear channel-wise dimensionality reduction\cite{lasby2024dynamicSparse}:
\begin{equation}
\tilde{\mathbf{F}} = \mathbf{W}_c \mathbf{F},
\end{equation}
where $\mathbf{W}_c \in \mathbb{R}^{C' \times C}$ is a learnable projection matrix with $C' \ll C$.

By projecting features into a compact subspace, the proposed module suppresses channel redundancy and constrains gradient variability, leading to more stable backpropagation. Moreover, reducing the dimensionality of privacy-sensitive representations alleviates noise amplification induced by differential privacy mechanisms, providing structured and stable inputs for subsequent clipping and aggregation.

\subsubsection{Gradient Sparsification}
To further enhance the stability of gradient updates, we apply sparsification to the reduced-dimensional gradients, thereby suppressing insignificant gradient components and promoting more stable backward propagation.

Specifically, the reduced-dimensional gradient associated with the reduced-dimensional feature representation $\tilde{\mathbf{F}}$ is defined as
\begin{equation}
\tilde{\mathbf{g}}=\frac{\partial \mathcal{L}}{\partial \tilde{\mathbf{F}}},
\label{eq:reduced_gradient}
\end{equation}
where $\mathcal{L}$ denotes the local training loss. We then apply magnitude-based sparsification to obtain the sparse gradient update as follows:
\begin{equation}
\hat{g}_i =
\begin{cases}
\tilde{g}_i, & \text{if } |\tilde{g}_i| \ge \tau_t, \\
0,           & \text{otherwise},
\end{cases}
\label{eq:gradient_sparsification}
\end{equation}
where $\tau_t$ is a sparsification threshold at training stage $t$, which decays dynamically over the course of training. A relatively larger threshold in the early training phase enforces stronger sparsity to stabilize update directions, while a smaller threshold in later stages preserves finer-grained gradient information to improve optimization capability.

\subsection{Adaptive Differentially Private Gradient Clipping}
In privacy-sensitive federated learning scenarios, differential privacy (DP) enforces strict limits on the influence of individual training samples by applying gradient clipping and injecting random noise during local training. However, conventional DP-SGD relies on a fixed clipping threshold, which is difficult to adapt to the dynamic variations of gradient updates in real-world distributed environments. Under the combined effects of Non-IID data distributions, heterogeneous client capabilities, and stage-wise training fluctuations, the gradient norms across clients often exhibit significant imbalance. As a result, a fixed clipping threshold may lead to over-clipping, which severely degrades the effective training signal, or under-clipping, which accelerates privacy budget consumption and amplifies noise perturbations. These issues ultimately impair model efficiency and final model performance. To address these challenges, we propose an adaptive differentially private gradient clipping mechanism based on gradient statistical information, enabling dynamic threshold adjustment to balance privacy protection and optimization stability.

\subsubsection{Local Gradient Clipping}
After LLDM-induced sparsification, the sparse feature-level gradient in (\ref{eq:gradient_sparsification}) is further backpropagated through the local network. Let $\hat{\mathbf{g}}_{i,k}^{(t)}$ denote the resulting gradient of the $k$-th trainable parameter tensor on client $i$ at round $t$. Each client then computes the global norm of its parameter gradients as
\begin{equation}
\|\hat{\nabla}_i^{(t)}\|_2 =
\left\|
\left[
\|\hat{\mathbf{g}}_{i,1}^{(t)}\|_2,\;
\|\hat{\mathbf{g}}_{i,2}^{(t)}\|_2,\;
\ldots,\;
\|\hat{\mathbf{g}}_{i,K}^{(t)}\|_2
\right]
\right\|_2,
\label{eq:global_grad_norm}
\end{equation}
where $\hat{\mathbf{g}}_{i,k}^{(t)}$ denotes the gradient of the $k$-th parameter tensor on client $i$. If the global gradient norm exceeds the current clipping threshold $C^{(t)}$, each gradient tensor is rescaled as
\begin{equation}
\bar{\mathbf{g}}_{i,k}^{(t)}
=
\hat{\mathbf{g}}_{i,k}^{(t)}
\cdot
\min\left(
1,\;
\frac{C^{(t)}}{\|\hat{\nabla}_i^{(t)}\|_2 + 10^{-6}}
\right),
\quad k=1,\ldots,K.
\label{eq:grad_clipping}
\end{equation}
This operation ensures that the overall update magnitude is bounded within a finite range, satisfying the sensitivity constraints required by differential privacy while effectively preventing training oscillations caused by abnormally large gradients.

\subsubsection{Adaptive Threshold Adjustment}
To address the limitation that fixed clipping thresholds cannot adapt to gradient variations across different training stages, we introduce an adaptive threshold adjustment strategy on the server side based on update norm statistics.

After obtaining the clipped gradients in (\ref{eq:grad_clipping}) and the privatized gradients later formalized in (\ref{eq:dp_noise}), each participating client performs local optimization starting from the broadcast global model $w_g^{(t)}$. Let the initialized local model on client $i$ be
\begin{equation}
w_{i,0}^{(t)} = w_g^{(t)}.
\label{eq:local_model_init}
\end{equation}
Using the privatized gradients, client $i$ then performs $E$ local optimization steps:
\begin{equation}
w_{i,e+1}^{(t)} = w_{i,e}^{(t)} - \eta \mathbf{g}_{i,e}^{\mathrm{dp},(t)},
\qquad e = 0,1,\ldots,E-1,
\label{eq:local_model_step}
\end{equation}
where $\eta$ is the local learning rate and $\mathbf{g}_{i,e}^{\mathrm{dp},(t)}$ denotes the privatized gradient used at local step $e$.

After completing local training, the accumulated model change on client $i$ during round $t$ is defined as
\begin{equation}
\Delta w_i^{(t)} = w_{i,E}^{(t)} - w_g^{(t)}.
\label{eq:model_delta}
\end{equation}
Accordingly, the resulting local model after client-side training can be written as
\begin{equation}
w_i^{(t+1)} = w_g^{(t)} + \Delta w_i^{(t)},
\label{eq:model_update}
\end{equation}
where $\Delta w_i^{(t)}$ is the model update uploaded by client $i$ to the server. The server then computes its $\ell_2$ norm as in per-update clipping based federated optimization~\cite{li2024updateclipping}:
\begin{equation}
s_i^{(t)} = \left\| \Delta w_i^{(t)} \right\|_2.
\label{eq:update_norm}
\end{equation}

The server then aggregates the update norms from all participating clients into a set $S^{(t)} = \{s_1^{(t)}, s_2^{(t)}, \ldots, s_m^{(t)}\}$ and uses the median of this set as the clipping threshold for the next communication round, following the general principle of norm-statistics-based adaptive clipping~\cite{fukami2025adaptiveclipping}:
\begin{equation}
C^{(t+1)} = \max \left( \mathrm{median}\left( S^{(t)} \right), C_{\min} \right).
\label{eq:adaptive_threshold}
\end{equation}

This adaptive mechanism dynamically captures the overall scale of uploaded client updates during training. Specifically, when most client updates are relatively small, the clipping threshold is automatically reduced to strengthen privacy protection; when Non-IID data distributions cause sudden increases in certain updates, the threshold is raised accordingly to prevent excessive clipping and preserve optimization stability.

\subsubsection{Differential Privacy Noise Injection}
After gradient clipping, let $\bar{\mathbf{g}}_{i}^{(t)}$ denote the clipped gradient update on client $i$. The perturbed gradient update is then computed as
\begin{equation}
\mathbf{g}_{i}^{\mathrm{dp},(t)} = \bar{\mathbf{g}}_{i}^{(t)} + \mathcal{N}\left(0, \sigma^2 C^{(t)2}\mathbf{I}\right),
\label{eq:dp_noise}
\end{equation}
where $\sigma$ denotes the noise scale corresponding to the privacy budget $\varepsilon$, and $C^{(t)}$ is the adaptive clipping threshold at communication round $t$.

In this work, a lightweight differential privacy budget (e.g., $\varepsilon = 8$) is adopted, which provides a reasonable trade-off between privacy protection and model utility while avoiding excessive degradation of training performance. By injecting noise after adaptive clipping, the sensitivity of each client gradient is effectively bounded, ensuring that the privacy guarantees remain valid under the standard Gaussian mechanism. The privatized gradient in (\ref{eq:dp_noise}) is not uploaded to the server directly. Instead, it is used in the client-side local optimization process in (\ref{eq:local_model_step}) to produce the uploaded model update $\Delta w_i^{(t)}$ in (\ref{eq:model_delta}), which is then used for server-side threshold adaptation in (\ref{eq:update_norm})--(\ref{eq:adaptive_threshold}).

Overall, the proposed adaptive differential privacy clipping mechanism achieves a principled balance among privacy preservation, model efficiency, and model accuracy by explicitly connecting gradient clipping, noise injection, client-side optimization, and server-side threshold adaptation.
\subsection{Constraint-Aware Robust Aggregation}
To alleviate the amplification of differential privacy noise and improve global convergence stability under client heterogeneity, we propose a constraint-aware robust aggregation strategy on the server side. Instead of directly constraining the aggregated model in the parameter space, the proposed method performs robustness modeling at the client-weight level. Specifically, the server first estimates a utility- and stability-aware nominal aggregation distribution according to local validation performance and update-stability statistics, and then conducts a constrained robust reweighting around this reference distribution. In this way, unreliable or noise-dominated client updates can be suppressed while excessive deviation from the nominal aggregation rule is avoided.

\subsubsection{Utility- and Stability-Aware Nominal Aggregation Weights}
At communication round $t$, each participating client $i \in \mathcal{S}_t$ uploads its local model update $\Delta w_i^{(t)}$ defined in (\ref{eq:model_delta}), together with a local validation performance score $q_i^{(t)}$ (measured by F1-score). To characterize the stability of local optimization, we use the update norm $s_i^{(t)}$ defined in (\ref{eq:update_norm}).

Since the raw statistics $q_i^{(t)}$ and $s_i^{(t)}$ may fluctuate considerably across communication rounds due to stochastic local optimization, heterogeneous data distributions, and DP perturbation, directly using them to determine aggregation weights can lead to unstable client reweighting~\cite{malekmohammadi2024noiseaware}. To reduce such short-term fluctuations, we maintain exponential moving averages as~\cite{morales2024ema}
\begin{equation}
\tilde{q}_i^{(t+1)} = \beta \tilde{q}_i^{(t)} + (1-\beta) q_i^{(t+1)},
\label{eq:ema_q}
\end{equation}
\begin{equation}
\tilde{s}_i^{(t+1)} = \beta \tilde{s}_i^{(t)} + (1-\beta) s_i^{(t+1)},
\label{eq:ema_s}
\end{equation}
where $\beta \in (0,1)$ is the smoothing factor. In this way, $\tilde{q}_i^{(t)}$ and $\tilde{s}_i^{(t)}$ provide more stable estimates of local utility and update stability by combining historical information with current observations, thereby reducing sensitivity to transient noise and round-wise variation~\cite{morales2024ema}.

Based on the smoothed utility and stability statistics, we define a utility-stability score
\begin{equation}
r_i^{(t)} = \frac{\tilde{q}_i^{(t)}}{\tilde{s}_i^{(t)} + \epsilon_0},
\label{eq:reliability_score}
\end{equation}
where $\epsilon_0 > 0$ is a small constant for numerical stability. The nominal aggregation weights are then obtained by
\begin{equation}
\alpha_i^{(t)} = \frac{\exp(r_i^{(t)})}{\sum_{j \in \mathcal{S}_t} \exp(r_j^{(t)})},
\label{eq:nominal_weight}
\end{equation}
where $\mathcal{S}_t$ denotes the set of participating clients in round $t$. Here, $\alpha^{(t)} = [\alpha_i^{(t)}]_{i \in \mathcal{S}_t}$ serves as the reference aggregation distribution for the current round.

The corresponding nominal aggregated update is
\begin{equation}
\bar{\Delta w}^{(t)} = \sum_{i \in \mathcal{S}_t} \alpha_i^{(t)} \Delta w_i^{(t)},
\label{eq:nominal_update}
\end{equation}
which serves as the nominal aggregation direction for subsequent deviation measurement and robust reweighting.

\subsubsection{Ambiguity Set Around the Nominal Aggregation Distribution}
Although utility- and stability-aware weighting alleviates part of the inconsistency among client updates, directly using $\alpha_i^{(t)}$ may still be vulnerable to severe Non-IID drift and differential privacy noise. To further enhance robustness, we construct an ambiguity set around the nominal aggregation distribution $\alpha^{(t)}$, following prior-distribution-centered ambiguity-set design in distributionally robust optimization~\cite{jiao2022distributed}.

For each participating client, we define an admissible deviation radius
\begin{equation}
\bar{p}_i^{(t)} = \rho \alpha_i^{(t)},
\label{eq:deviation_radius}
\end{equation}
where $\rho \in (0,1]$ controls the allowable deviation from the nominal aggregation weight. The ambiguity set is then defined as
\begin{equation}
\mathcal{P}_t =
\left\{
p \in \mathbb{R}_+^{|\mathcal{S}_t|}
\;\middle|\;
\begin{aligned}
&\sum_{i \in \mathcal{S}_t} p_i = 1,\\
&|p_i - \alpha_i^{(t)}| \le \bar{p}_i^{(t)}, \quad \forall\, i \in \mathcal{S}_t,\\
&\sum_{i \in \mathcal{S}_t}
\frac{|p_i - \alpha_i^{(t)}|}{\bar{p}_i^{(t)} + \epsilon_0}
\le \Gamma_t
\end{aligned}
\right\},
\label{eq:ambiguity_set}
\end{equation}
where $\Gamma_t$ is the robustness budget at round $t$. A smaller $\Gamma_t$ keeps the aggregation closer to the nominal aggregation distribution, whereas a larger $\Gamma_t$ allows stronger suppression of unreliable or noise-dominated client contributions~\cite{jiao2022distributed}.

\subsubsection{Direction-Aware Robust Reweighting}
Although the nominal weights $\alpha_i^{(t)}$ already incorporate utility and update-stability information, they may still be sensitive to directional inconsistency caused by DP perturbation and heterogeneous local data distributions. To further improve aggregation robustness, we measure the deviation of each client update from the nominal aggregated direction:
\begin{equation}
d_i^{(t)} = \left\| \Delta w_i^{(t)} - \bar{\Delta w}^{(t)} \right\|_2.
\label{eq:direction_deviation}
\end{equation}
A larger $d_i^{(t)}$ indicates that the update from client $i$ is less consistent with the dominant aggregation direction and is therefore more likely to destabilize the global optimization process.

Based on these deviation statistics, the ideal robust aggregation weight vector can be formulated as the solution to
\begin{equation}
p^{(t),*} = \arg\min_{p \in \mathcal{P}_t} \sum_{i \in \mathcal{S}_t} p_i d_i^{(t)}.
\label{eq:robust_reweighting_obj}
\end{equation}
This formulation searches for a weight vector that preserves the utility- and stability-aware nominal weighting structure while suppressing updates with large directional deviation.

To avoid solving (\ref{eq:robust_reweighting_obj}) exactly with an additional inner optimization loop, we adopt a lightweight one-step projected descent update to construct an approximate robust weight vector:
\begin{equation}
\tilde{p}^{(t)} = \Pi_{\mathcal{P}_t}
\left(
\alpha^{(t)} - \eta_p d^{(t)}
\right),
\label{eq:robust_weight_update}
\end{equation}
where $d^{(t)} = [d_i^{(t)}]_{i \in \mathcal{S}_t}$, $\eta_p$ is the step size, and $\Pi_{\mathcal{P}_t}(\cdot)$ denotes the projection onto the ambiguity set $\mathcal{P}_t$. In practice, we use $\tilde{p}^{(t)}$ as the robust aggregation weight vector in round $t$.

Following recent studies on federated aggregation, the server forms a weighted combination of client updates and then applies the aggregated update to the current global model~\cite{nanayakkara2024globalagg}:
\begin{equation}
\Delta w_{\mathrm{rob}}^{(t)} = \sum_{i \in \mathcal{S}_t} \tilde{p}_i^{(t)} \Delta w_i^{(t)},
\label{eq:robust_update}
\end{equation}
and the global model is updated as
\begin{equation}
w_g^{(t+1)} = w_g^{(t)} + \Delta w_{\mathrm{rob}}^{(t)}.
\label{eq:robust_global_model}
\end{equation}

This robust aggregation mechanism does not require an additional inner optimization loop. Instead, it performs an efficient constrained reweighting around the nominal aggregation distribution according to client-level utility-stability and directional deviation statistics. As a result, the server can suppress unstable or noise-dominated updates while preserving the efficiency of standard round-based federated optimization. The complete training procedure of the proposed method is summarized in Algorithm~\ref{alg:fed-dimdp}.
\begin{algorithm}[t]
\caption{FedDimDP: Adaptive Differentially Private Federated Learning}
\label{alg:fed-dimdp}
\begin{algorithmic}[1]
\REQUIRE Initial global model $w_g^{(0)}$, total communication rounds $T$, privacy budget $(\varepsilon,\delta)$, initial clipping threshold $C^{(0)}$
\ENSURE Final global model $w_g^{(T)}$

\FOR{$t = 0, 1, \dots, T-1$}
    \STATE Server broadcasts the current global model $w_g^{(t)}$ to all selected clients
    \FORALL{clients $i \in \mathcal{S}_t$ \textbf{in parallel}}
        \STATE Receive global model $w_g^{(t)}$
        \STATE Perform local training with LLDM
        \STATE Compute local parameter gradients and apply clipping according to (\ref{eq:grad_clipping})
        \STATE Inject Gaussian noise to obtain privatized local gradients according to (\ref{eq:dp_noise})
        \STATE Initialize the local model according to (\ref{eq:local_model_init})
        \STATE Perform local optimization using the privatized gradients according to (\ref{eq:local_model_step})
        \STATE Compute the uploaded model update $\Delta w_i^{(t)}$ according to (\ref{eq:model_delta})
        \STATE Evaluate local validation performance $q_i^{(t)}$
        \STATE Upload $\Delta w_i^{(t)}$ and $q_i^{(t)}$ to the server
    \ENDFOR

    \STATE Server computes update norms $s_i^{(t)}$ according to (\ref{eq:update_norm}) for all $i \in \mathcal{S}_t$
    \STATE Update adaptive clipping threshold according to (\ref{eq:adaptive_threshold})
    \STATE Update smoothed statistics according to (\ref{eq:ema_q})--(\ref{eq:ema_s})
    \STATE Compute utility-stability scores $r_i^{(t)}$ and nominal weights $\alpha_i^{(t)}$ according to (\ref{eq:reliability_score})--(\ref{eq:nominal_weight})
    \STATE Compute nominal aggregated update $\bar{\Delta w}^{(t)}$ according to (\ref{eq:nominal_update})
    \STATE Compute directional deviations $d_i^{(t)}$ according to (\ref{eq:direction_deviation})
    \STATE Construct the ambiguity set $\mathcal{P}_t$ according to (\ref{eq:ambiguity_set})
    \STATE Construct approximate robust aggregation weights $\tilde{p}^{(t)}$ via (\ref{eq:robust_weight_update})
    \STATE Compute robust aggregated update $\Delta w_{\mathrm{rob}}^{(t)}$ according to (\ref{eq:robust_update})
    \STATE Update global model $w_g^{(t+1)} \leftarrow w_g^{(t)} + \Delta w_{\mathrm{rob}}^{(t)}$
\ENDFOR

\STATE \textbf{return} $w_g^{(T)}$
\end{algorithmic}
\end{algorithm}
\section{Experimental Setup}
This section evaluates the proposed adaptive differentially private federated learning framework under Non-IID data distributions. We conduct overall performance comparisons and ablation studies on the CIFAR-10, SVHN, and STL-10 datasets to examine the effectiveness of key components, including the local dimensionality reduction module (LLDM), adaptive gradient clipping, and constraint-aware robust aggregation.

In all experiments, we consider $N = 10$ clients, each performing $E = 1$ local epoch per communication round. For fair comparison, all methods are implemented using the same CNN backbone for image classification, consisting of two convolutional layers followed by two fully connected layers. Optimization is performed using stochastic gradient descent (SGD) with learning rate $0.01$ and a local batch size of $64$. All experiments were conducted on a Linux server equipped with an NVIDIA A30 GPU (24 GB VRAM). The models were implemented using PyTorch 2.1 and Python 3.9.

\subsection{Datasets and Non-IID Data Partitioning}
We evaluate our method on CIFAR-10, SVHN, and STL-10. CIFAR-10 contains 60,000 $32 \times 32$ RGB images from 10 classes. SVHN is a real-world digit recognition dataset with 73,257 training images and 26,032 test images. STL-10 is a more challenging natural image classification benchmark with 10 classes, including 5,000 labeled training images, 8,000 test images, and 100,000 unlabeled images. Since our focus is supervised federated learning under differential privacy, we use the labeled training split and the standard test split for evaluation.

To simulate heterogeneous client data distributions, we partition all datasets across $N = 10$ clients under Non-IID settings. For CIFAR-10 and STL-10, we adopt Dirichlet-based label-skew partitioning, where class proportions are sampled from $\mathrm{Dirichlet}(\alpha)$ with $\alpha \in \{0.3, 0.1\}$ to simulate moderate and severe Non-IID settings, respectively. For SVHN, samples are grouped by class and unevenly distributed among clients, resulting in highly imbalanced local label distributions.

\section{Results and Analysis}
\subsection{Overall Performance Comparison}
\begin{table}[t]
\caption{Classification performance on CIFAR-10, SVHN, and STL-10 under differential privacy.}
\label{tab:main_results}
\centering
\begin{tabular}{lcccccc}
\hline
Method & \multicolumn{2}{c}{CIFAR-10} & \multicolumn{2}{c}{SVHN} & \multicolumn{2}{c}{STL-10} \\
\cline{2-7}
 & Acc & F1 & Acc & F1 & Acc & F1 \\
\hline
DP-FedSAM~\cite{bao2024asafedsam}  & 0.742 & 0.742 & 0.786 & 0.860 & 0.577 & 0.581 \\
DP-ACDN~\cite{xue2024adaptiveNoiseTIFS}    & 0.707 & 0.708 & 0.777 & 0.833 & 0.623 & 0.624 \\
FedACG~\cite{chen2024dpflnoniidtkde}     & 0.684 & 0.682 & 0.880 & 0.874 & 0.583 & 0.585 \\
AWDP-FL~\cite{wang2024multiprivacy}    & 0.564 & 0.558 & 0.798 & 0.769 & 0.553 & 0.555 \\
FedSA~\cite{zhang2024lightweightdpfl}      & 0.522 & 0.493 & 0.769 & 0.762 & 0.519 & 0.527 \\
FedDimDP   & \textbf{0.811} & \textbf{0.809} & \textbf{0.897} & \textbf{0.890} & \textbf{0.652} & \textbf{0.650} \\
\hline
\end{tabular}
\end{table}

Table~\ref{tab:main_results} reports the classification performance on CIFAR-10, SVHN, and STL-10 under unified differential privacy settings. We observe that different baselines exhibit distinct strengths, yet their performance is still limited by the coupled challenges of privacy perturbation and data heterogeneity. DP-FedSAM, which incorporates sharpness-aware minimization to enhance generalization under DP, achieves relatively strong results on CIFAR-10 (Acc/F1: 0.742/0.742), but its advantage becomes less pronounced when gradient signals are heavily distorted by clipping and noise. DP-ACDN further mitigates privacy noise through adaptive communication denoising and achieves the strongest baseline performance on STL-10 (0.623/0.624), suggesting that denoising is beneficial in relatively challenging visual settings; however, its gains on CIFAR-10 and SVHN remain moderate (0.707/0.708 and 0.777/0.833), indicating that denoising alone cannot fully resolve stage-wise gradient-scale variation and cross-client inconsistency under Non-IID data. FedACG dynamically adjusts clipping thresholds to stabilize DP-SGD and shows competitive performance on SVHN (0.880/0.874), yet its results on CIFAR-10 and STL-10 (0.684/0.682 and 0.583/0.585) indicate that clipping adaptation without explicit robustness control is still vulnerable to heterogeneous update drift. AWDP-FL introduces adaptive privacy weighting across heterogeneous clients, but its relatively weak performance across all three datasets implies that overly conservative privacy weighting may suppress useful optimization signals. FedSA, as a FedAvg-style DP baseline with fixed-variance Gaussian noise, yields the weakest performance overall, further highlighting the sensitivity of fixed DP configurations to non-stationary gradient distributions.

In contrast, the proposed FedDimDP consistently achieves the best performance on all three datasets (CIFAR-10 Acc/F1: 0.811/0.809; SVHN Acc/F1: 0.897/0.890; STL-10 Acc/F1: 0.652/0.650). On CIFAR-10, FedDimDP improves accuracy by 6.84\% over the strongest baseline DP-FedSAM (0.8108 vs. 0.7424), demonstrating superior robustness under label-skew heterogeneity and DP perturbation. On SVHN, it surpasses the best competing method FedACG by 1.76\% in accuracy (0.8974 vs. 0.8798), indicating stronger stability on real-world data distributions. On STL-10, it further outperforms the strongest baseline DP-ACDN by 2.88\% in accuracy (0.6522 vs. 0.6234), confirming its effectiveness on a more challenging visual classification benchmark. These gains can be attributed to the joint effect of: (i) the lightweight local dimensionality reduction module (LLDM), which regularizes intermediate representations and suppresses noise amplification, (ii) adaptive DP clipping that aligns the threshold with historical update statistics to reduce over-/under-clipping, and (iii) constraint-aware robust aggregation that down-weights unreliable updates while stabilizing the aggregated direction against DP-induced perturbation and Non-IID drift. Overall, the results confirm that FedDimDP achieves a more favorable utility--stability trade-off than existing DP-FL baselines under heterogeneous clients and privacy noise.
\subsection{Ablation Study}
\begin{table}[t]
\centering
\caption{Ablation results on CIFAR-10, SVHN, and STL-10 datasets.}
\label{tab:ablation}
\begin{tabular}{lcc|cc|cc}
\toprule
\multirow{2}{*}{Method} & \multicolumn{2}{c|}{CIFAR-10} & \multicolumn{2}{c|}{SVHN} & \multicolumn{2}{c}{STL-10} \\
 & Acc & F1 & Acc & F1 & Acc & F1 \\
\midrule
w/o-CRA     & 0.722 & 0.600 & 0.863 & 0.848 & 0.597 & 0.587 \\
w/o-ADPC    & 0.794 & 0.791 & 0.878 & 0.867 & 0.650 & 0.648 \\
w/o-LLDM    & 0.736 & 0.736 & 0.817 & 0.802 & 0.618 & 0.610 \\
\midrule
FedDimDP    & \textbf{0.811} & \textbf{0.809} & \textbf{0.897} & \textbf{0.890} & \textbf{0.652} & \textbf{0.650} \\
\bottomrule
\end{tabular}
\end{table}

To assess the contribution of each component, we conduct ablation studies on CIFAR-10, SVHN, and STL-10, as reported in Table~\ref{tab:ablation}. Removing constraint-aware robust aggregation (w/o-CRA) causes the most significant performance degradation on CIFAR-10 and STL-10, indicating that the proposed robust aggregation is critical for stabilizing global convergence under heterogeneous data distributions. Disabling adaptive differentially private clipping (w/o-ADPC) leads to a relatively moderate performance drop across all three datasets, showing that adaptive clipping helps maintain a better balance between privacy perturbation and optimization stability. Removing the lightweight local dimensionality reduction module (w/o-LLDM) consistently degrades performance on all datasets and results in the largest drop on SVHN, highlighting its important role in stabilizing local training and improving representation robustness. Overall, these results confirm that robust aggregation, adaptive clipping, and local dimensionality reduction contribute complementarily to the final performance, with w/o-CRA and w/o-LLDM showing the most noticeable degradation under differential privacy and Non-IID settings.

\section{Conclusion}
In this paper, we proposed FedDimDP, an adaptive differentially private federated learning framework that improves model efficiency under Non-IID data and DP perturbations by combining the lightweight local dimensionality reduction module, adaptive DP clipping, and constraint-aware robust aggregation. Experiments on CIFAR-10, SVHN, and STL-10 demonstrate consistent gains in both accuracy and F1 compared with representative DP-FL baselines. As future work, we will extend FedDimDP to decentralized and asynchronous federated settings, and further study its privacy--utility trade-off under more diverse vision backbones and real-world system heterogeneity.
\bibliographystyle{IEEEtran}
\bibliography{SMC2026}
\end{document}